\documentclass[runningheads]{llncs}
\usepackage[T1]{fontenc}
\usepackage{graphicx}
\usepackage{hyperref}
\usepackage{color}

\usepackage{siunitx}
\usepackage{xcolor}
\usepackage{amsmath}
\usepackage{amssymb}
\usepackage{booktabs}
\usepackage{multirow}
\usepackage[misc]{ifsym}
\usepackage{xcolor}
\newcommand\crule[3][black]{\textcolor{#1}{\rule{#2}{#3}}}

\definecolor{rightkidney}{RGB}{255,232,0}
\definecolor{liver}{RGB}{110,15,202}
\definecolor{stomach}{RGB}{251,14,255}
\definecolor{leftkidney}{RGB}{63,182,73}
\definecolor{esophagus}{RGB}{0,114,255}
\definecolor{spleen}{RGB}{210,2,0}
\definecolor{pancreas}{RGB}{46,175,181}
\definecolor{aorta}{RGB}{170,107,0}
\definecolor{venacava}{RGB}{254,161,68}
\definecolor{gall}{RGB}{0,252,124}
\definecolor{lag}{RGB}{107,0,208}
\definecolor{rag}{RGB}{62,27,203}
\definecolor{portal_vein}{RGB}{0,132,32}

\begin{document}
\title{Unsupervised 3D registration through optimization-guided cyclical self-training}
\titlerunning{Unsupervised 3D registration through cyclical self-training}
%
\author{Alexander Bigalke\inst{1}\textsuperscript{(\Letter)}\and
Lasse Hansen\inst{2}\and
Tony C. W. Mok\inst{3}\and
Mattias P. Heinrich\inst{1}}

\authorrunning{A. Bigalke et al.}
%
\institute{Institute of Medical Informatics, University of L\"ubeck, L\"ubeck, Germany \email{\{alexander.bigalke,mattias.heinrich\}@uni-luebeck.de} \and
EchoScout GmbH, L\"ubeck, Germany \and
DAMO Academy, Alibaba Group
}
%
\maketitle              
\begin{abstract}
State-of-the-art deep learning-based registration methods employ three different learning strategies: supervised learning, which requires costly manual annotations, unsupervised learning, which heavily relies on hand-crafted similarity metrics designed by domain experts, or learning from synthetic data, which introduces a domain shift.
To overcome the limitations of these strategies, we propose a novel self-supervised learning paradigm for unsupervised registration, relying on self-training.
Our idea is based on two key insights.
Feature-based differentiable optimizers 1) perform reasonable registration even from random features and 2) stabilize the training of the preceding feature extraction network on noisy labels.
Consequently, we propose cyclical self-training, where pseudo labels are initialized as the displacement fields inferred from random features and cyclically updated based on more and more expressive features from the learning feature extractor, yielding a self-reinforcement effect.
We evaluate the method for abdomen and lung registration, consistently surpassing metric-based supervision and outperforming diverse state-of-the-art competitors.
Source code is available at \url{https://github.com/multimodallearning/reg-cyclical-self-train}.

\end{abstract}

\section{Introduction}
Medical image registration is a fundamental task in medical imaging with applications ranging from multi-modal data fusion to temporal data analysis.
In recent years, deep learning has advanced learning-based registration methods \cite{haskins2020deep}, which achieve competitive performances at low runtimes and thus constitute a promising alternative to accurate but slow classical optimization methods.
A decisive factor in successfully training deep learning-based methods is the choice of a suitable strategy to supervise the learning process.
In the literature, there exist three different learning strategies.
The first is supervised learning based on manual annotations such as landmark correspondences \cite{hansen2021deep} or semantic labels \cite{hu2018weakly}.
However, manual annotations are costly and may introduce a label bias \cite{balakrishnan2019voxelmorph}.
Alternatively, a second strategy employs synthetic deformation fields to generate image pairs with precisely known displacement fields \cite{eppenhof2018pulmonary}.
However, this introduces a domain gap between synthetic training and real test pairs, limiting the performance at inference time.
Elaborated deformation techniques can reduce the gap but require strong domain knowledge, are tailored to specific problems, and do not generalize across tasks.
The third widely used training strategy is unsupervised metric-based learning, maximizing a similarity metric between fixed and warped moving images, e.g.~implemented in \cite{balakrishnan2019voxelmorph,mok2020large}.
Popular metrics include normalized cross-correlation \cite{sarvaiya2009image} and MIND \cite{heinrich2012mind}.
However, the success of this strategy strongly depends on the specific hand-crafted metric, and the performance of the trained deep learning models is often inferior to a classical optimization-based counterpart.
Considering the deficiencies of the above training techniques, in this work, we introduce a novel learning strategy for unsupervised registration based on the concept of self-training.

Self-training is a widespread training strategy for semi-supervised learning \cite{xie2020self} and domain adaptation \cite{zou2018unsupervised}.
The core idea is to pre-train a network on available labeled data and subsequently apply the model to the unlabeled data to generate so-called pseudo labels.
Afterwards, one alternates between re-training the model on the union of labeled and pseudo-labeled data and updating the pseudo labels with the current model.
This general concept was successfully adapted to diverse tasks and settings, with methods in medical context primarily focusing on segmentation \cite{hang2020local,perone2019unsupervised}.
These methods resort to a special form of self-training, the Mean Teacher paradigm \cite{tarvainen2017mean}, where pseudo labels are continuously provided by a teacher model, representing a temporal ensemble of the learning network.
A persistent problem of classical and Mean Teacher-based self-training is the inherent noise of the pseudo labels, which can severely hamper the learning process.
As a remedy, some works aim to filter reliable pseudo labels based on model uncertainty \cite{yu2019uncertainty}.
Only recently, the Mean Teacher was adapted to the registration problem, tackling domain adaptation \cite{bigalke2022adapting} or complementing metric-based supervision for adaptive regularization weighting \cite{xu2022double}.
Contrary to these methods, we introduce self-training for registration in a fully unsupervised setting, with pseudo labels as the single source of supervision.

\textbf{Contributions.}
We introduce a novel learning paradigm for unsupervised registration by adapting the concept of self-training to the problem.
This involves two principal challenges.
First, labeled data for the pre-training stage is unavailable, raising the question of how to generate initial pseudo labels.
Second, as a general problem in self-training, the negative impact of noise in the pseudo labels needs to be mitigated.
In our pursuit to overcome these challenges, we made two decisive observations (see Fig.~\ref{fig:cum_dsc}) when exploring a combination of deep learning-based feature extraction with differentiable optimization algorithms for the displacement prediction, such as \cite{hansen2021deep,siebert2022learn}.
First, we found that feature-based optimizers predict reasonable displacement fields and improve the initial registration even when applied to the output of random feature networks (orange line in Fig.~\ref{fig:cum_dsc}).
We attribute this feature to the inductive bias of deep neural networks, which extract somewhat useful features even with random weights \cite{cao2022random}.
These predicted displacements thus constitute meaningful initial pseudo labels, solving the first problem and leaving us with the second problem to overcome the noise in the labels.
In this context, we made the second observation that the intrinsic regularizing capacity of the optimizers stabilizes the learning from noisy labels.
Specifically, training the feature extractor on our initial pseudo labels yielded registrations surpassing the accuracy of the noisy labels used for training (green, red, purple, brown, and magenta lines in Fig.~\ref{fig:cum_dsc}).
Consequently, we propose a cyclical self-training scheme, alternating between training the feature extractor and updating the pseudo labels.
As such, our novel learning paradigm does not require costly manual annotations, prevents the domain shift of synthetic deformations, and is independent of hand-crafted similarity metrics.
Moreover, our method significantly differs from previous uncertainty-based pseudo label filtering strategies since it implicitly overcomes the negative impact of noisy labels by combining deep feature learning with regularizing differentiable optimization.
We evaluate the method for CT abdomen registration and keypoint-based lung registration, demonstrating substantial improvements over diverse state-of-the-art comparison methods.

\section{Methods}
\subsection{Problem setup}
Given a data pair $(\boldsymbol{F},\boldsymbol{M})$ of a fixed and a moving image as input, registration aims at finding a displacement field $\boldsymbol{\varphi}$ that spatially aligns $\boldsymbol{M}$ to $\boldsymbol{F}$.
We address the task in an unsupervised setting, where training data $\mathcal{T}=\{(\boldsymbol{F}_i,\boldsymbol{M}_i)\}_{i=1}^{|\mathcal{T}|}$ consists of $|\mathcal{T}|$ unlabeled data pairs.
Given the training data, we aim to learn a function $f$ with parameters $\boldsymbol{\theta}_f$, (partially) represented by a deep network, which predicts displacement fields as $\hat{\boldsymbol{\varphi}}=f(\boldsymbol{F},\boldsymbol{M};\boldsymbol{\theta}_f)$.

\subsection{Cyclical self-training}
\begin{figure}[t]
\centering
\includegraphics[width=\textwidth]{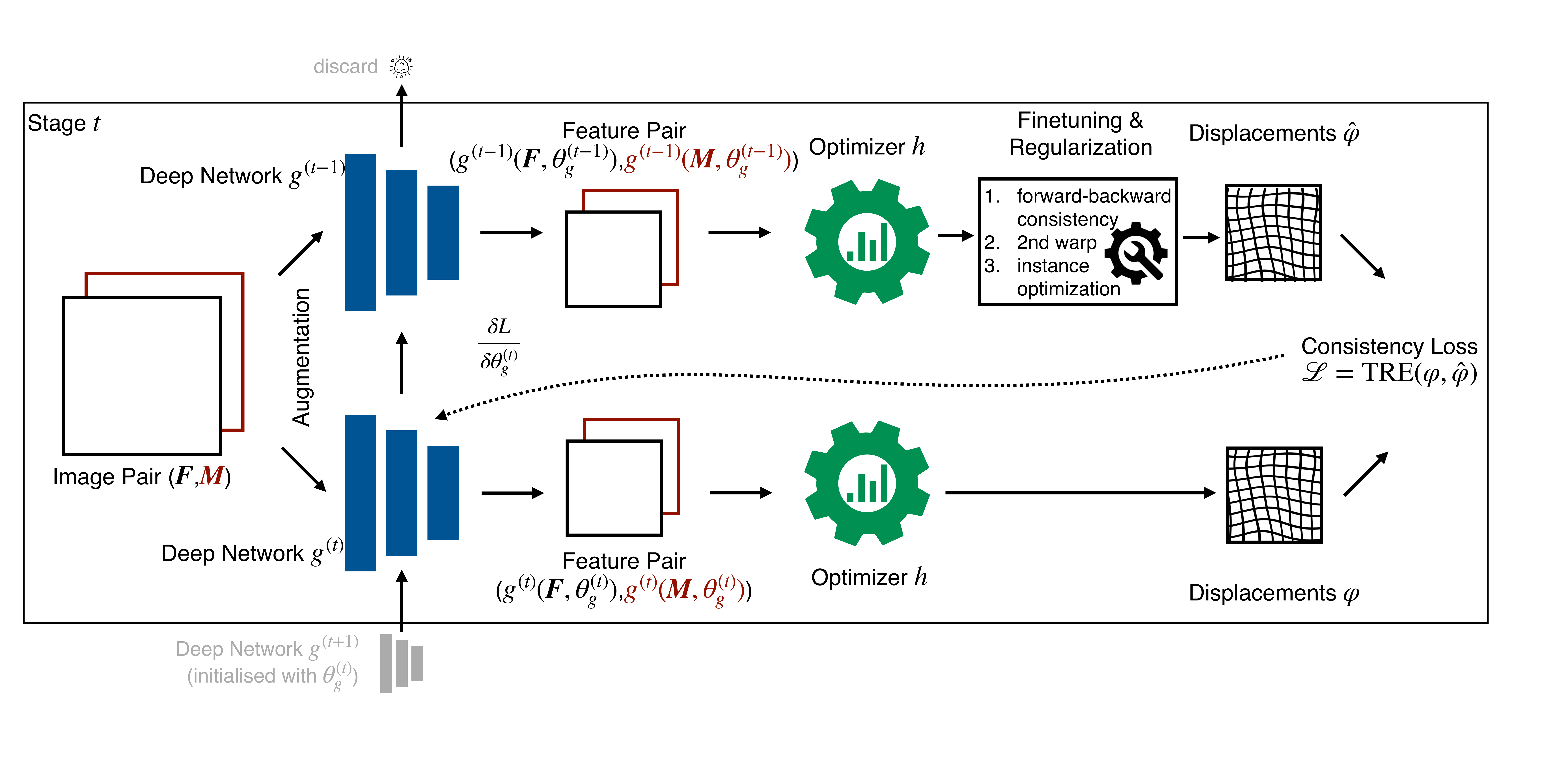}
\caption{Overview of the proposed cyclical self-training paradigm for unsupervised registration. The underlying registration pipeline comprises a deep network for feature extraction $g$ and a differentiable optimizer $h$ to predict the displacements.
At stage $t$, we supervise the training of the network $g^{(t)}$ with pseudo labels generated based on the features from the network $g^{(t-1)}$ from the previous stage. For optimal feature learning, the pseudo displacements from the optimizer are further refined and regularized.}
\label{fig:overview}
\end{figure}

We propose to solve the above problem with a cyclical self-training strategy visualized in Fig.~\ref{fig:overview}.
While existing self-training methods assume the availability of some labeled data, annotations are unavailable in our unsupervised setting.
To overcome this issue and generate an initial set of pseudo labels for the first stage of self-training, we parameterize the function $f$ as the combination of a deep neural network $g$ for feature extraction with a non-learnable but differentiable feature-based optimization algorithm $h$ for displacement prediction, i.e.
\begin{equation}
    f(\boldsymbol{F},\boldsymbol{M};\boldsymbol{\theta}_f)=h(g(\boldsymbol{F},\boldsymbol{M};\boldsymbol{\theta}_g))
\end{equation}
The approach is based on our empirical observation that a suitable optimization algorithm $h$ can predict reasonable initial displacement fields $\hat{\boldsymbol{\varphi}}^{(0)}$ from random features provided by a network $g^{(0)}$ with random initialization $\boldsymbol{\theta}_g^{(0)}$, which is in line with recent studies on the inductive bias of CNNs \cite{cao2022random}.
We leverage these predicted displacements as pseudo labels to supervise the first stage of self-training, where the parameters of the feature extractor with different initialization $\boldsymbol{\theta}_g^{(1)}$ are optimized by minimizing the loss
\begin{equation}
    \mathcal{L}(\boldsymbol{\theta}^{(1)}_g;\mathcal{T})=\frac{1}{|\mathcal{T}|}\sum_i\mathrm{TRE}\left(h\left(g\left(\boldsymbol{F}_i,\boldsymbol{M}_i;\boldsymbol{\theta}^{(1)}_g\right)\right),\hat{\boldsymbol{\varphi}}_i^{(0)}\right)
\end{equation}
with $\rm{TRE}(\hat{\boldsymbol{\varphi}}_i^{(1)},\hat{\boldsymbol{\varphi}}_i^{(0)})$ denoting the mean over the element-wise target registration error between the displacement fields $\hat{\boldsymbol{\varphi}}_i^{(1)}$ and $\hat{\boldsymbol{\varphi}}_i^{(0)}$.

A critical problem of this basic setup is that the network might overfit the initial pseudo labels and learn to reproduce random features.
Therefore, in the spirit of recent techniques from contrastive learning \cite{chen2021exploring}, we propose to improve the efficacy of feature learning by incorporating asymmetries into the learning and pseudo label streams at two levels.
First, we apply different random augmentations to the input pairs in both streams.
Second, we augment the pseudo label stream with additional (non-differentiable) fine-tuning and regularization steps after the optimizer to improve the pseudo displacement fields (see Sec.~\ref{sec:registration_framework} for details).
As demonstrated in our ablation experiments (Fig.~\ref{fig:cum_dsc}, Tab.~\ref{tab:ablation}), both strategies improve feature learning and strengthen the self-improvement effect.

Once the first stage of self-training has converged, we repeat the process $T$ times.
Specifically, at stage $t$, we generate refined pseudo labels with the trained network $g^{(t-1)}$ from the previous stage, initialize the learning network $g^{(t)}$ with the weights from $g^{(t-1)}$ and perform a warm restart on the learning rate to escape potential local minima from the previous stage.

\subsection{Registration framework}
\label{sec:registration_framework}
Our proposed self-training scheme is a flexible, modular framework, agnostic to the input modality and the specific implementation of feature extractor $g$ and optimizer $h$.
This section describes our specific design choices for $g$ and $h$ for image and point cloud registration, with the former being our main focus.

\subsubsection{Image registration.}
To extract features from 3D input volumes, we implement $g$ a standard 3D CNN with six convolution layers with kernel sizes $3\times 3\times 3$ and 32, 64, or 128 channels.
Each convolution is followed by BatchNorm and ReLU, and every second convolution contains a stride of 2, yielding a downsampling factor of 8.
The outputs for both images are mapped to 16-dimensional features using a $1\times1\times1$ convolution and fed into a correlation layer \cite{sun2018pwc} that captures 125 discrete displacements.

As the optimizer, we adapt the coupled convex optimization for learning-based 3D registration from \cite{siebert2022learn}, which, given fixed and moving features, infers a displacement field that minimizes a combined objective of smoothness and feature dissimilarity.
Our proposed refinement strategy in the pseudo label stream comprises three ingredients.
1) Forward-backward consistency additionally computes the reverse displacement field ($\boldsymbol{F}$ to $\boldsymbol{M}$) and then iteratively minimizes the discrepancy between both fields.
2) For a second warp, the moving image is warped with the inferred displacement field before repeating all previous steps.
3) Iterative instance optimization finetunes the final displacement field with Adam by jointly minimizing regularization cost and feature dissimilarity.
For the latter, we use the CNN features after the second convolution block and map them with a $1\times1\times1$ convolution to 16 channels.
We apply the same refinement steps at test time.
Moreover, we propose to leverage the difference between network-predicted and finetuned displacements to estimate the difficulty of the training samples.
Consequently, we apply a weighted batch sampling at training that increases the probability of using less difficult registration pairs with a higher agreement between both fields.
We rank all training pairs and use a sigmoid function with arguments ranging linearly from -5 to 5 for the weighted random sampler.

\subsubsection{Point cloud registration.}
For point cloud registration, we implement the feature extractor as a graph CNN and rely on sparse loopy belief propagation for differentiable optimization, as introduced in \cite{hansen2021deep}.

\section{Experiments}
\subsection{Experimental setup}
\subsubsection{Datasets.}
We conduct our main experiments for inter-patient abdomen CT registration using the corresponding dataset of the Learn2Reg (L2R) Challenge\footnote{\url{https://learn2reg.grand-challenge.org/}} \cite{hering2022learn2reg}.
The dataset contains 30 abdominal 3D CT scans of different patients with 13 manually labeled anatomical structures of strongly varying sizes.
The original image data and labels are from \cite{xu2016evaluation}.
As part of L2R, they were affinely pre-registered into a canonical space and resampled to identical voxel resolutions (\SI{2}{mm}) and spatial dimensions ($192\times 160\times \SI{256}{vx}$).
Following the data split of L2R, we use 20 scans (190 pairs) for training and the remaining 10 scans (45 pairs) for evaluation.
Hence, data split and preprocessing are consistent with compared previous works \cite{hansen2021deep,yan2022sam}. 
As metrics, we report the mean Dice overlap (DSC) between the semantic labels and the standard deviation of the logarithmic Jacobian determinant (SDlogJ).

We perform a second experiment for inhale-to-exhale lung CT registration on the DIR-Lab COPDGene dataset\footnote{\url{https://med.emory.edu/departments/radiation-oncology/research-laboratories/deformable-image-registration/downloads-and-reference-data/copdgene.html}} \cite{castillo2013reference}, which comprises 10 such scan pairs.
For each pair, 300 expert-annotated landmark correspondences are available for evaluation.
We pre-process all scans in multiple steps: 1) resampling to $1.75\times 1.00\times \SI{1.25}{mm}$ for exhale and $1.75\times 1.25\times \SI{1.75}{mm}$ for inhale, 2)
cropping with fixed-size bounding boxes ($192\times 192\times \SI{208}{vx}$), centered around automatically generated lung masks, 3) affine pre-registration, aligning the lung masks.
Since we focus on keypoint-based registration of the lung CTs, we follow \cite{hansen2021deep} and extract distinctive keypoints from the CTs using the F\"orstner algorithm with non-maximum suppression, yielding around 1k points in the fixed and 2k points in the moving cloud.
In our experiments, we perform 5-fold cross-validation, with each fold comprising eight data pairs for training and two for testing.
We report the target registration error (TRE) at the landmarks and the SDlogJ as metrics. 

\subsubsection{Implementation details.}
We implement all methods in Pytorch and optimize network parameters with the Adam optimizer.
For abdomen registration, we train for $T=8$ stages, each stage comprising 1000 iterations with a batch size of 2.
The learning rate follows a cosine annealing warm restart schedule, decaying from $10^{-3}$ to {$10^{-5}$} at each stage.
Hyper-parameters were set based on the DSC on three cases from the training set.
For lung registration, the model converged after $T=5$ stages of 60 epochs with batch size 4, with an initial learning rate of 0.001, decreased by a factor of 10 at epochs 40 and 52.
Here, hyper-parameters were adopted from \cite{hansen2021deep}.
For both datasets, training requires 90-\SI{100}{min} and \SI{8}{GB} on an RTX2080, and input augmentations consist of random affine transformations.

\subsection{Results}
\subsubsection{Abdomen.}
First, we analyze our method in several ablation experiments.
In Fig.~\ref{fig:cum_dsc}, we visualize the performance of our method on a subset of classes over several cycles of self-training.
We observe consistent improvements over the stages, particularly pronounced at early stages while the performance converges later on.
\begin{table}[t]
\begin{minipage}{0.5\textwidth}
\centering
\includegraphics[width=\textwidth]{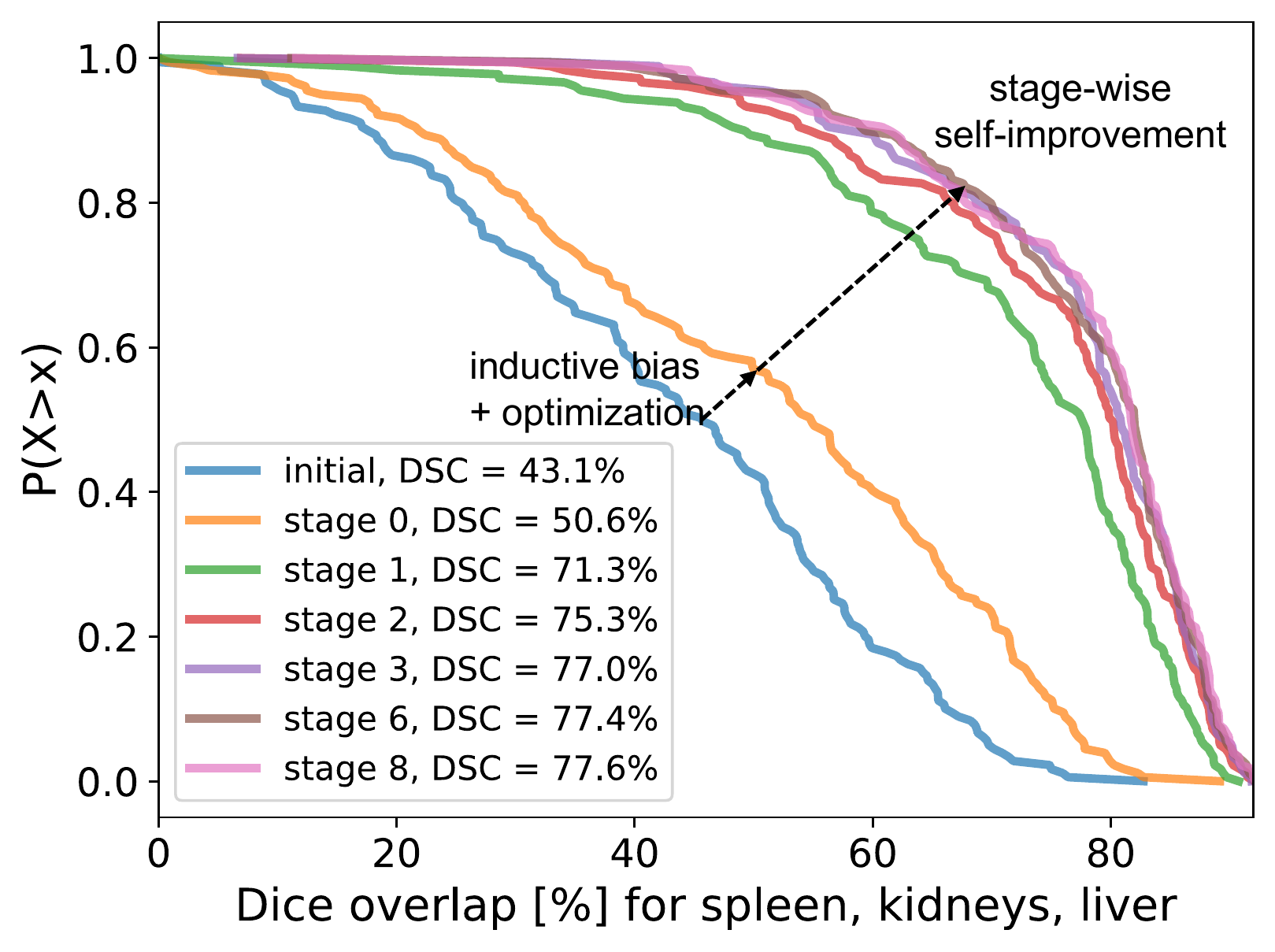}
\renewcommand\tablename{Fig.}
\renewcommand{\thetable}{2}
\caption{``Opposite'' cumulative distribution of Dice overlaps for Abdomen CT registration after different stages of self-training.}
\label{fig:cum_dsc}
\end{minipage}
\hfill
\begin{minipage}{0.45\textwidth}
\renewcommand{\thetable}{1}
\caption{Ablation study for abdomen CT registration.}
\label{tab:ablation}
\begin{tabular}{lcc}
\toprule
Method &  DSC & SDlogJ\\
\midrule
prealign & 25.9 & -\\
\midrule
w/o input augm. &  48.8 & .129\\
w/o PL refinement & 48.8 & .200\\ 
w/o weighted sampling &  50.1 & .147\\
ours & $\boldsymbol{51.1}$ & .146\\
\midrule
1 warp w/o Adam &  38.6 & $\boldsymbol{.061}$\\
1 warp w/ Adam &  49.6 & .119\\
2 warps w/o Adam & 41.1 & .088\\
2 warps w/Adam (ours) &  $\boldsymbol{51.1}$ & .146\\
\bottomrule
\end{tabular}
\end{minipage}
\end{table}
This highlights the self-reinforcing effect achieved through alternating pseudo label updates and network training.
\begin{figure}[b!]
\centering
\includegraphics[width=\textwidth]{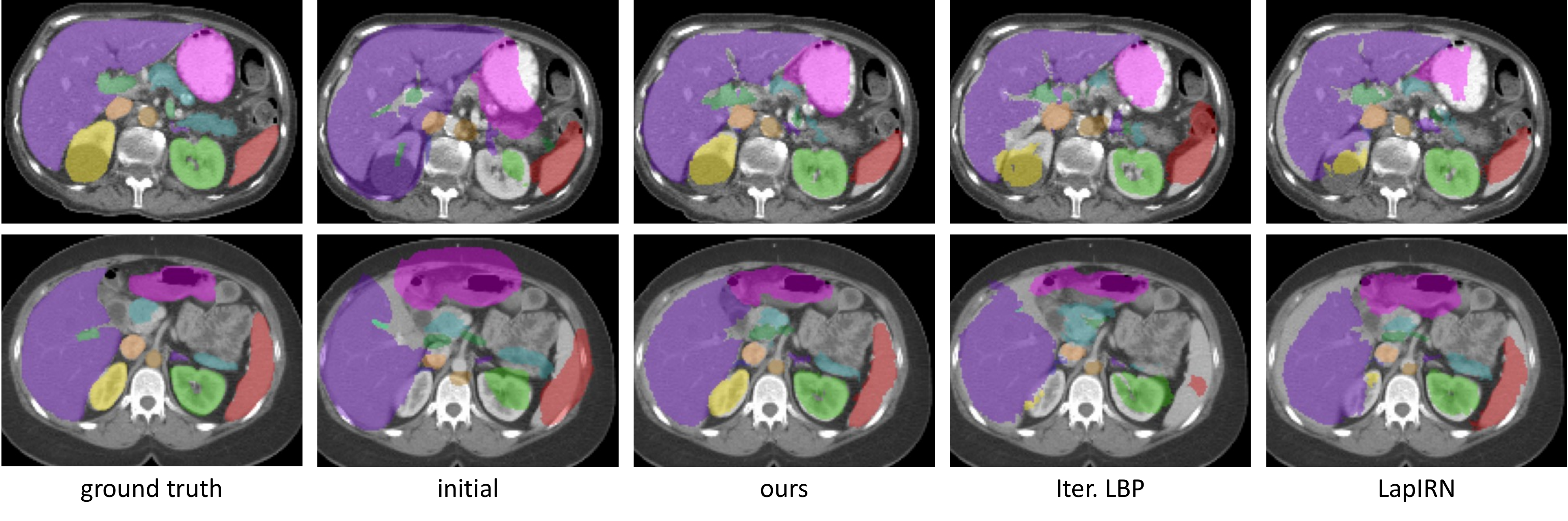}
\renewcommand{\thefigure}{3}
\caption{Qualitative results of selected methods on two cases of the     Abdomen CT dataset (axial view).
    We show overlays of the warped segmentation labels with the fixed scan: liver \crule[liver]{5pt}{5pt}, stomach \crule[stomach]{5pt}{5pt}, left kidney \crule[leftkidney]{5pt}{5pt}, right kidney \crule[rightkidney]{5pt}{5pt}, spleen \crule[spleen]{5pt}{5pt}, gall bladder \crule[gall]{5pt}{5pt}, esophagus \crule[esophagus]{5pt}{5pt}, pancreas \crule[pancreas]{5pt}{5pt}, aorta \crule[aorta]{5pt}{5pt}, inferior vena cava \crule[venacava]{5pt}{5pt}, portal vein \crule[portal_vein]{5pt}{5pt}, left \crule[lag]{5pt}{5pt}/right \crule[rag]{5pt}{5pt} adrenal gland.
}
\label{fig:qual_results}
\end{figure}
In the upper part of Tab.~\ref{tab:ablation}, we verify the efficacy of incorporating asymmetries (input augmentations, finetuning of pseudo labels) into both streams and weighted sampling.
The results confirm the importance of each component to reach optimal performance.
In the lower part of Tab.~\ref{tab:ablation}, we evaluate our final model under different test configurations, highlighting the improvements through a second warp and Adam finetuning.

\begin{figure}[h]
\centering
	\begin{center}
        \includegraphics[width=\linewidth]{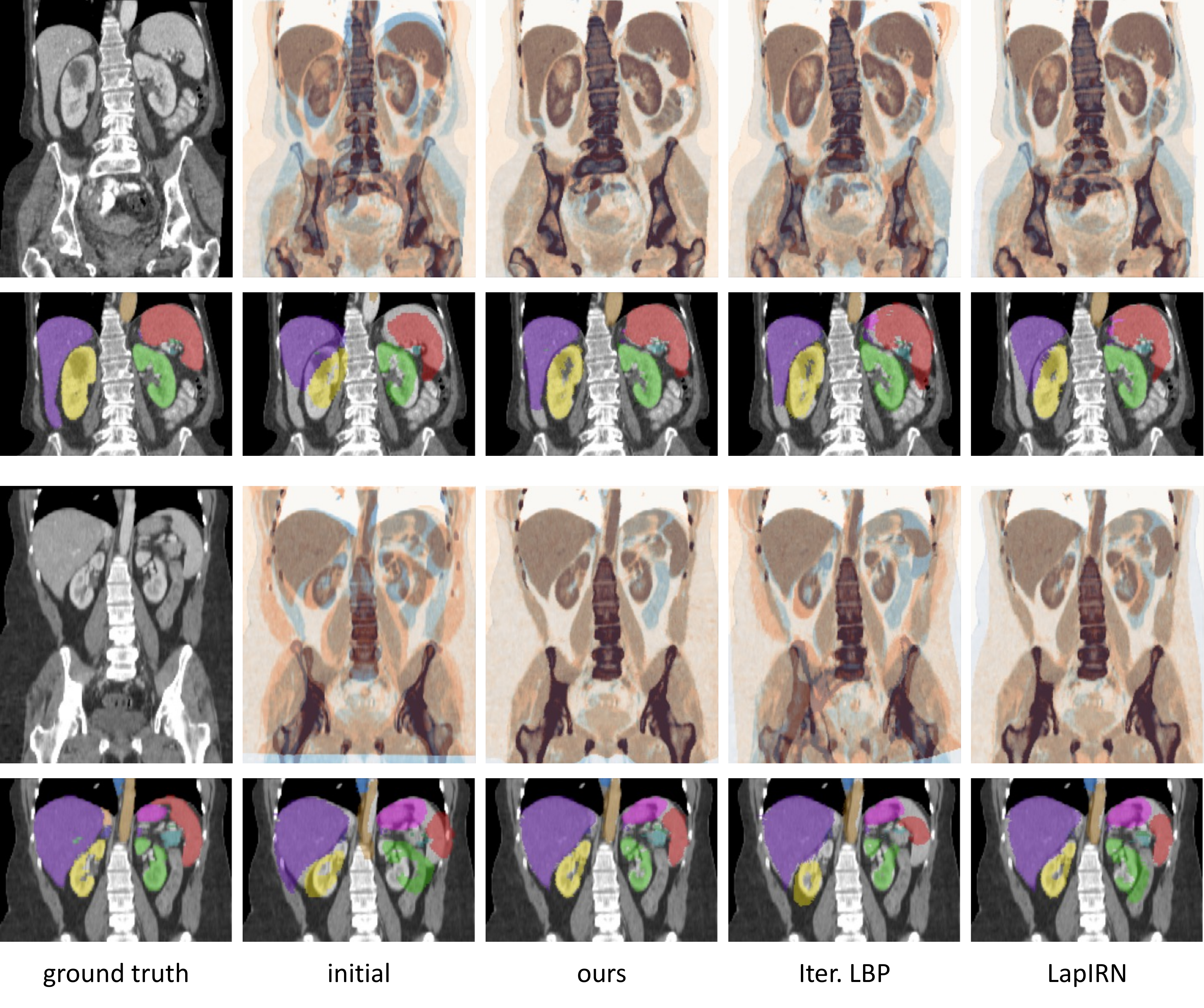}
	\end{center}
	\caption{Qualitative results on two cases of the Abdomen CT dataset from coronal view.
	Rows 1 and 3 display overlaid CT slices of fixed (blue) and warped moving (orange) scans.
	Colors add up to grayscale when the scans are perfectly aligned.
	Rows 2 and 4 show overlays of the warped segmentation labels with the fixed scan: liver \crule[liver]{5pt}{5pt}, stomach \crule[stomach]{5pt}{5pt}, left kidney \crule[leftkidney]{5pt}{5pt}, right kidney \crule[rightkidney]{5pt}{5pt}, spleen \crule[spleen]{5pt}{5pt}, gall bladder \crule[gall]{5pt}{5pt}, esophagus \crule[esophagus]{5pt}{5pt}, pancreas \crule[pancreas]{5pt}{5pt}, aorta \crule[aorta]{5pt}{5pt}, inferior vena cava \crule[venacava]{5pt}{5pt}, portal vein \crule[portal_vein]{5pt}{5pt}, left \crule[lag]{5pt}{5pt}/right \crule[rag]{5pt}{5pt} adrenal gland.}
    \label{fig:supp:abdomen}
\end{figure}

Next, we compare our method to a comprehensive set of state-of-the-art unsupervised methods, including classical algorithms \cite{avants2008symmetric,hansen2021revisiting,heinrich2013mrf} and deep learning-based approaches, trained with MIND \cite{balakrishnan2019voxelmorph,heinrich2019closing}/NCC \cite{mok2020large} supervision or contrastive learning \cite{yan2022sam}.
The results are collected from \cite{hansen2021deep} and \cite{yan2022sam}.
Moreover, we train our own registration framework with metric-based supervision (MIND \cite{heinrich2012mind}, NCC \cite{sarvaiya2009image}) to directly verify the advantage of our self-training strategy.
Results are shown in Tab.~\ref{tab:abdomen_sota_results}, Fig.~\ref{fig:qual_results}, and Supp., Fig.~\ref{fig:supp:abdomen}.
Our method substantially outperforms all comparison methods in terms of DSC (statistical significance is confirmed by a Wilcoxon-signed rank test with p<0.0001 for all competitors with public code, which excludes SAME) and sets a new state-of-the-art accuracy of 51.1\% DSC.
This highlights the advantages of our new learning paradigm over previous unsupervised strategies.
Meanwhile, the smoothness of predicted displacement fields (SDlogJ) is comparable with most unsupervised deep learning-based methods \cite{balakrishnan2019voxelmorph,heinrich2019closing} and superior to MIND- and NCC-supervision.
\begin{table}[t]
\renewcommand{\thetable}{2}
\begin{minipage}[t]{0.5\textwidth}
\caption{Results for unsupervised abdomen CT registration.}
\label{tab:abdomen_sota_results}
\setlength{\tabcolsep}{1pt}
\begin{tabular}{@{}lccc@{}}
    \toprule
    Method & Dice [\%] & SDlogJ & Time [s]\\
    \midrule
    pre-aligned & 25.9 & -\\
    \midrule
    Adam \cite{hansen2021revisiting} & 36.6 & $\boldsymbol{.080}$ & 1.6\\
    Iter.~LBP \cite{hansen2021revisiting} & 40.1 & .093 & 0.6\\
    ANTs (SyN) \cite{avants2008symmetric} & 28.4 & N/A & 74.3\\ 
    DEEDS \cite{heinrich2013mrf} & 46.5 & N/A & 45.4\\
    \midrule
    VoxelMorph \cite{balakrishnan2019voxelmorph} & 35.4 & .134 & $\boldsymbol{0.2}$\\
    PDD \cite{heinrich2019closing} & 41.5 & .129 & 1.4\\
    LapIRN \cite{mok2020large}& 42.4 & .089 & 3.8\\
    SAME \cite{yan2022sam} & 49.8 & N/A & 1.2\\
    \midrule
    MIND sup. & 47.7 & .237 & 1.2\\
    NCC sup. & 48.1 & .299 & 1.2\\
    \midrule
    ours & $\boldsymbol{51.1}$ & .146 & 1.2\\
    \bottomrule
\end{tabular}
\end{minipage}
\hfill
\begin{minipage}[t]{0.45\textwidth}
\renewcommand{\thetable}{3}
\setlength{\tabcolsep}{1pt}
\caption{Results for lung CT registration on the COPD dataset.}
\label{tab:lung_sota}
  \begin{tabular}{@{}lcc@{}}
    \toprule
    Method & TRE [mm] & SDlogJ \\
    \midrule
    initial (pre-aligned) & 11.99 & -\\
    \midrule
    VoxelMorph \cite{balakrishnan2019voxelmorph} & 7.98 & N/A\\
    LapIRN \cite{mok2020large} & 4.99 & N/A\\
    PDD \cite{heinrich2019closing} & 2.16 & N/A\\
    \midrule
    rigid deform. & 2.98 & .037\\
    rnd.~field deform. & 3.19 & .035\\
    metric sup. \cite{wu2020pointpwc} & 6.79 & .042\\
    landmark sup. \cite{hansen2021deep} & 2.27 & .036\\
    \midrule
    ours & $\boldsymbol{1.93}$ & $\boldsymbol{.033}$\\
    \bottomrule
\end{tabular}
\end{minipage}
\end{table}

\subsubsection{Lung.}
\begin{figure}[t]
\centering
	\begin{center}
        \includegraphics[width=\linewidth]{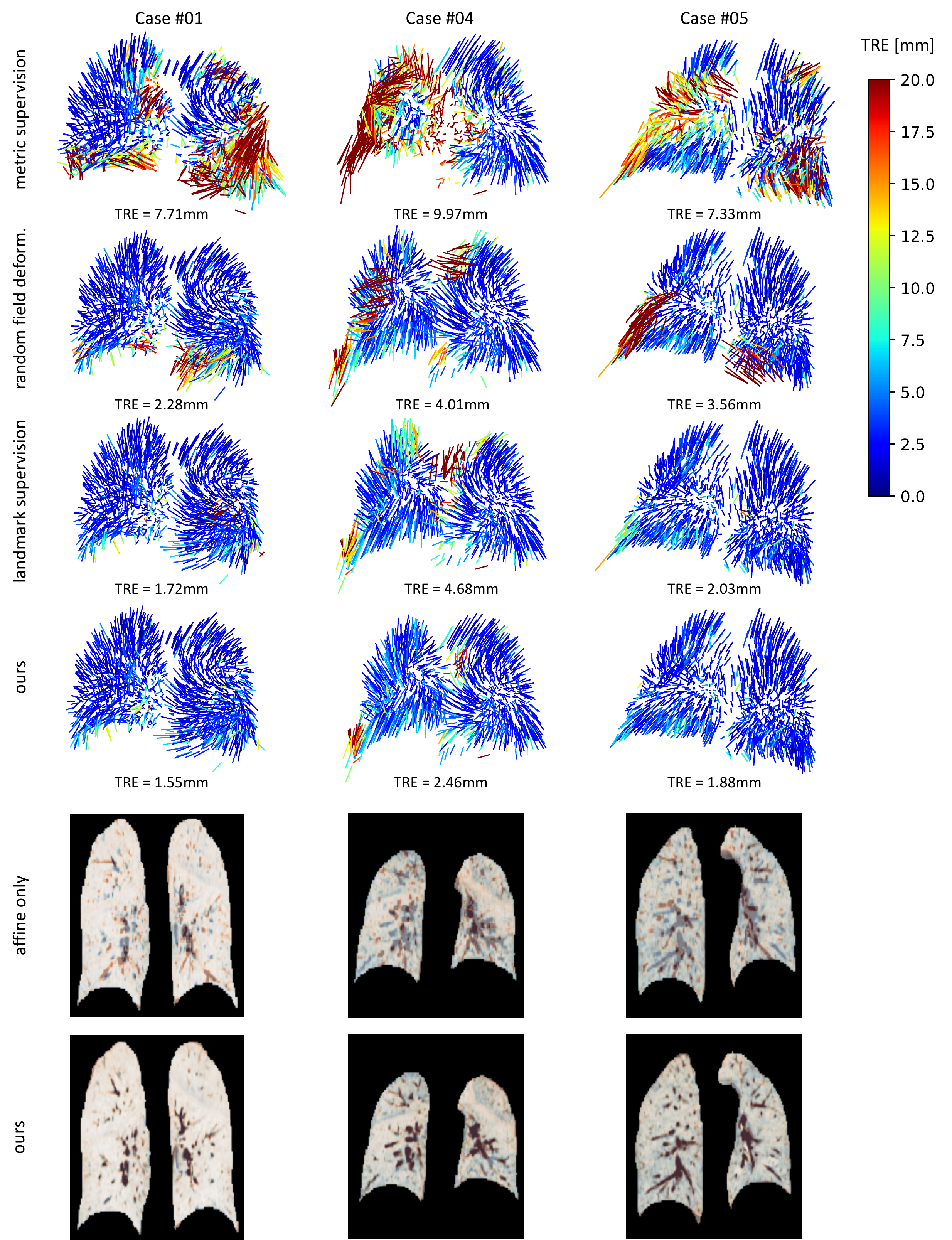}
	\end{center}
	\caption{Qualitative results for keypoint-based registration on the COPD dataset.
	The first four rows show the displacement vectors at the F\"orstner keypoints predicted by different methods.
	Colors represent the target registration error (TRE) of the individual vectors clamped to \SI{20}{mm}.
	The last two rows display overlaid CT slices before and after registration by our method.
	Fixed and moving scans are shown in blue and orange, adding up to grayscale when perfectly aligned.}
    \label{fig:supp:lung}
\end{figure}

For point cloud-based lung registration, we compare our cyclical self-training strategy to three alternative learning strategies: supervision with manually annotated landmark correspondences as in \cite{hansen2021deep}, metric-based supervision with Chamfer distance and local Laplacian penalties as in \cite{wu2020pointpwc}, and training on synthetic rigid/random field deformations.
All strategies are implemented for the same baseline registration model from \cite{hansen2021deep}.
Moreover, we report the performance of three unsupervised image-based deep learning methods \cite{balakrishnan2019voxelmorph,heinrich2019closing,mok2020large} trained with MIND supervision.
Results are shown in Tab.~\ref{tab:lung_sota}, demonstrating the superiority of our self-training strategy over all competing learning strategies and the reported image-based SOTA methods.
Qualitative results of the experiment are shown in Supp., Fig.~\ref{fig:supp:lung}, demonstrating accurate and smooth displacements, as also confirmed by low values of SDlogJ.

\section{Conclusion}
We introduced a novel cyclical self-training paradigm for unsupervised registration.
To this end, we developed a modular registration pipeline of a deep feature extraction network coupled with a differentiable optimizer, stabilizing learning from noisy pseudo labels through regularization and iterative, cyclical refinement.
That way, our method avoids pitfalls of popular metric supervision (NCC, MIND), which relies on shallow features or image intensities and is prone to noise and local minima.
By contrast, our supervision through optimization-refined and -regularized pseudo labels promotes learning task-specific features that are more robust to noise, and our cyclical learning strategy gradually improves the expressiveness of features to avoid local minima.
In our experiments, we demonstrated the efficacy and flexibility of our approach, which outperformed the competing state-of-the-art methods and learning strategies for dense image-based abdomen and point cloud-based lung registration.
In summary, we did not only present the first fully unsupervised self-training scheme but also a new perspective on unsupervised learning-based registration.
In particular, we consider our strategy complementary to existing techniques (metric-based and contrastive learning), opening up the potential for combined training schemes in future work.

\paragraph{Acknowledgement.} We gratefully acknowledge the financial support by the \linebreak Federal Ministry for Economic Affairs and Climate Action of Germany \linebreak (FKZ: 01MK20012B) and by the Federal Ministry for Education and Research of Germany (FKZ: 01KL2008).

%
%
\bibliographystyle{splncs04}
\bibliography{paper1171-bibliography}

\begin{thebibliography}{10}
\providecommand{\url}[1]{\texttt{#1}}
\providecommand{\urlprefix}{URL }
\providecommand{\doi}[1]{https://doi.org/#1}

\bibitem{avants2008symmetric}
Avants, B.B., Epstein, C.L., Grossman, M., Gee, J.C.: Symmetric diffeomorphic
  image registration with cross-correlation: evaluating automated labeling of
  elderly and neurodegenerative brain. Medical image analysis  \textbf{12}(1),
  26--41 (2008)

\bibitem{balakrishnan2019voxelmorph}
Balakrishnan, G., Zhao, A., Sabuncu, M.R., Guttag, J., Dalca, A.V.: Voxelmorph:
  a learning framework for deformable medical image registration. IEEE
  transactions on medical imaging  \textbf{38}(8),  1788--1800 (2019)

\bibitem{bigalke2022adapting}
Bigalke, A., Hansen, L., Heinrich, M.P.: Adapting the mean teacher for
  keypoint-based lung registration under geometric domain shifts. In:
  International Conference on Medical Image Computing and Computer-Assisted
  Intervention. pp. 280--290. Springer (2022)

\bibitem{cao2022random}
Cao, Y.H., Wu, J.: A random cnn sees objects: One inductive bias of cnn and its
  applications. In: Proceedings Of The AAAI Conference On Artificial
  Intelligence. vol.~36, pp. 194--202 (2022)

\bibitem{castillo2013reference}
Castillo, R., Castillo, E., Fuentes, D., Ahmad, M., Wood, A.M., Ludwig, M.S.,
  Guerrero, T.: A reference dataset for deformable image registration spatial
  accuracy evaluation using the copdgene study archive. Physics in Medicine \&
  Biology  \textbf{58}(9), ~2861 (2013)

\bibitem{chen2021exploring}
Chen, X., He, K.: Exploring simple siamese representation learning. In:
  Proceedings of the IEEE/CVF conference on computer vision and pattern
  recognition. pp. 15750--15758 (2021)

\bibitem{eppenhof2018pulmonary}
Eppenhof, K.A., Pluim, J.P.: Pulmonary ct registration through supervised
  learning with convolutional neural networks. IEEE transactions on medical
  imaging  \textbf{38}(5),  1097--1105 (2018)

\bibitem{hang2020local}
Hang, W., Feng, W., Liang, S., Yu, L., Wang, Q., Choi, K.S., Qin, J.: Local and
  global structure-aware entropy regularized mean teacher model for 3d left
  atrium segmentation. In: International Conference on Medical Image Computing
  and Computer-Assisted Intervention. pp. 562--571. Springer (2020)

\bibitem{hansen2021deep}
Hansen, L., Heinrich, M.P.: Deep learning based geometric registration for
  medical images: How accurate can we get without visual features? In:
  International Conference on Information Processing in Medical Imaging. pp.
  18--30. Springer (2021)

\bibitem{hansen2021revisiting}
Hansen, L., Heinrich, M.P.: Revisiting iterative highly efficient optimisation
  schemes in medical image registration. In: International Conference on
  Medical Image Computing and Computer-Assisted Intervention. pp. 203--212.
  Springer (2021)

\bibitem{haskins2020deep}
Haskins, G., Kruger, U., Yan, P.: Deep learning in medical image registration:
  a survey. Machine Vision and Applications  \textbf{31}(1),  1--18 (2020)

\bibitem{heinrich2019closing}
Heinrich, M.P.: Closing the gap between deep and conventional image
  registration using probabilistic dense displacement networks. In:
  International Conference on Medical Image Computing and Computer-Assisted
  Intervention. pp. 50--58. Springer (2019)

\bibitem{heinrich2012mind}
Heinrich, M.P., Jenkinson, M., Bhushan, M., Matin, T., Gleeson, F.V., Brady,
  M., Schnabel, J.A.: Mind: Modality independent neighbourhood descriptor for
  multi-modal deformable registration. Medical image analysis  \textbf{16}(7),
  1423--1435 (2012)

\bibitem{heinrich2013mrf}
Heinrich, M.P., Jenkinson, M., Brady, M., Schnabel, J.A.: Mrf-based deformable
  registration and ventilation estimation of lung ct. IEEE transactions on
  medical imaging  \textbf{32}(7),  1239--1248 (2013)

\bibitem{hering2022learn2reg}
Hering, A., Hansen, L., Mok, T.C., Chung, A.C., Siebert, H., H{\"a}ger, S.,
  Lange, A., Kuckertz, S., Heldmann, S., Shao, W., et~al.: Learn2reg:
  comprehensive multi-task medical image registration challenge, dataset and
  evaluation in the era of deep learning. IEEE Transactions on Medical Imaging
  (2022)

\bibitem{hu2018weakly}
Hu, Y., Modat, M., Gibson, E., Li, W., Ghavami, N., Bonmati, E., Wang, G.,
  Bandula, S., Moore, C.M., Emberton, M., et~al.: Weakly-supervised
  convolutional neural networks for multimodal image registration. Medical
  image analysis  \textbf{49},  1--13 (2018)

\bibitem{mok2020large}
Mok, T.C., Chung, A.: Large deformation diffeomorphic image registration with
  laplacian pyramid networks. In: International Conference on Medical Image
  Computing and Computer-Assisted Intervention. pp. 211--221. Springer (2020)

\bibitem{perone2019unsupervised}
Perone, C.S., Ballester, P., Barros, R.C., Cohen-Adad, J.: Unsupervised domain
  adaptation for medical imaging segmentation with self-ensembling. NeuroImage
  \textbf{194},  1--11 (2019)

\bibitem{sarvaiya2009image}
Sarvaiya, J.N., Patnaik, S., Bombaywala, S.: Image registration by template
  matching using normalized cross-correlation. In: 2009 international
  conference on advances in computing, control, and telecommunication
  technologies. pp. 819--822. IEEE (2009)

\bibitem{siebert2022learn}
Siebert, H., Heinrich, M.P.: Learn to fuse input features for large-deformation
  registration with differentiable convex-discrete optimisation. In:
  International Workshop on Biomedical Image Registration. pp. 119--123.
  Springer (2022)

\bibitem{sun2018pwc}
Sun, D., Yang, X., Liu, M.Y., Kautz, J.: Pwc-net: Cnns for optical flow using
  pyramid, warping, and cost volume. In: Proceedings of the IEEE conference on
  computer vision and pattern recognition. pp. 8934--8943 (2018)

\bibitem{tarvainen2017mean}
Tarvainen, A., Valpola, H.: Mean teachers are better role models:
  Weight-averaged consistency targets improve semi-supervised deep learning
  results. Advances in neural information processing systems  \textbf{30}
  (2017)

\bibitem{wu2020pointpwc}
Wu, W., Wang, Z.Y., Li, Z., Liu, W., Fuxin, L.: Pointpwc-net: Cost volume on
  point clouds for (self-) supervised scene flow estimation. In: European
  Conference on Computer Vision. pp. 88--107. Springer (2020)

\bibitem{xie2020self}
Xie, Q., Luong, M.T., Hovy, E., Le, Q.V.: Self-training with noisy student
  improves imagenet classification. In: Proceedings of the IEEE/CVF conference
  on computer vision and pattern recognition. pp. 10687--10698 (2020)

\bibitem{xu2022double}
Xu, Z., Luo, J., Lu, D., Yan, J., Frisken, S., Jagadeesan, J., Wells~III, W.M.,
  Li, X., Zheng, Y., Tong, R.K.y.: Double-uncertainty guided spatial and
  temporal consistency regularization weighting for learning-based abdominal
  registration. In: Medical Image Computing and Computer Assisted Intervention
  -- MICCAI 2022. pp. 14--24. Springer (2022)

\bibitem{xu2016evaluation}
Xu, Z., Lee, C.P., Heinrich, M.P., Modat, M., Rueckert, D., Ourselin, S.,
  Abramson, R.G., Landman, B.A.: Evaluation of six registration methods for the
  human abdomen on clinically acquired ct. IEEE Transactions on Biomedical
  Engineering  \textbf{63}(8),  1563--1572 (2016)

\bibitem{yan2022sam}
Yan, K., Cai, J., Jin, D., Miao, S., Guo, D., Harrison, A.P., Tang, Y., Xiao,
  J., Lu, J., Lu, L.: Sam: Self-supervised learning of pixel-wise anatomical
  embeddings in radiological images. IEEE Transactions on Medical Imaging
  \textbf{41}(10),  2658--2669 (2022)

\bibitem{yu2019uncertainty}
Yu, L., Wang, S., Li, X., Fu, C.W., Heng, P.A.: Uncertainty-aware
  self-ensembling model for semi-supervised 3d left atrium segmentation. In:
  International Conference on Medical Image Computing and Computer-Assisted
  Intervention. pp. 605--613. Springer (2019)

\bibitem{zou2018unsupervised}
Zou, Y., Yu, Z., Kumar, B., Wang, J.: Unsupervised domain adaptation for
  semantic segmentation via class-balanced self-training. In: Proceedings of
  the European conference on computer vision (ECCV). pp. 289--305 (2018)

\end{thebibliography}

\newpage
\section*{Supplementary Material}

\end{document}